\def\year{2022}\relax
\documentclass[letterpaper]{article} 
\usepackage{aaai22}  
\usepackage{times}  
\usepackage{helvet}  
\usepackage{courier}  
\usepackage[hyphens]{url}  
\usepackage{graphicx} 
\usepackage{natbib}  
\usepackage{caption} 
\DeclareCaptionStyle{ruled}{labelfont=normalfont,labelsep=colon,strut=off} 
\usepackage{algorithm}
\usepackage{algorithmic}

\usepackage{newfloat}
\usepackage{listings}
\floatstyle{ruled}
\newfloat{listing}{tb}{lst}{}
\floatname{listing}{Listing}
\usepackage{amsmath}
\usepackage{amssymb}
\usepackage{amsfonts,bm}
\usepackage{kotex} 
\usepackage{multirow}
\usepackage{booktabs}
\usepackage{url}
\usepackage{subfigure}
\usepackage{adjustbox}
\usepackage[normalem]{ulem}
\usepackage{color}

\newcommand{\eg}{\textit{e}.\textit{g}.}

\def\1{\bm{1}}

\title{Saliency Grafting: Innocuous Attribution-Guided Mixup\\ with Calibrated Label Mixing}
\author{
    Joonhyung Park, \textsuperscript{\rm 1} 
    June Yong Yang, \textsuperscript{\rm 1}  
    Jinwoo Shin, \textsuperscript{\rm 1} 
    Sung Ju Hwang, \textsuperscript{\rm 1,2} 
    Eunho Yang \textsuperscript{\rm 1,2}
    
}
\affiliations{
    \textsuperscript{\rm 1}Korea Advanced Institute of Science and Technology (KAIST) \\
    \textsuperscript{\rm 2}AITRICS \\
    \{deepjoon, laoconeth, jinwoos, sjhwang82, eunhoy\}@kaist.ac.kr

}

\usepackage{bibentry}

\begin{document}

\maketitle

\begin{abstract}
The Mixup scheme suggests mixing a pair of samples to create an augmented training sample 
and has gained considerable attention recently for improving the generalizability of neural networks. A straightforward and widely used extension of Mixup is to combine with regional dropout-like methods: removing random patches from a sample and replacing it with the features from another sample. Albeit their simplicity and effectiveness, these methods are prone to create harmful samples due to their randomness. To address this issue, `maximum saliency' strategies were recently proposed: they select only the most informative features to prevent such a phenomenon. 
However, they now suffer from 
lack of sample diversification as they always deterministically select regions with maximum saliency, injecting bias into the augmented data. 
In this paper, we present 
a novel, yet simple Mixup-variant that captures the best of both worlds. Our idea is two-fold. By stochastically sampling the features and ‘grafting’ them onto another sample, our method effectively generates diverse yet meaningful samples. Its second ingredient is to produce the label of the grafted sample by mixing the labels in a saliency-calibrated fashion, which rectifies supervision misguidance introduced by the random sampling procedure. Our experiments under CIFAR, Tiny-ImageNet, and ImageNet datasets show that our scheme outperforms the current state-of-the-art augmentation strategies not only in terms of classification accuracy, but is also superior in coping under stress conditions such as data scarcity and object occlusion.

\end{abstract}


\section{Introduction}


Modern deep neural networks (DNNs) have achieved unprecedented success in various
computer vision tasks, 
\eg, image classification \cite{he2016deep}, generation \cite{brock2018large} and segmentation \cite{he2017mask}.
However, due to their over-parameterized nature, DNNs require an immense amount of training data to generalize well for test data. Otherwise, DNNs are predisposed to memorize the training samples and exhibit lackluster performance on the unseen data - in other words, incur overfitting.

Acquiring a sufficient amount of data for a given task is not always possible as it consumes valuable manpower and budget. One common approach to combat such data scarcity is \emph{data augmentation}, which aims to enlarge the effective size of a dataset by producing virtual samples from the training data through means such as injecting noise \cite{amodei2016deepspeech} or cropping out regions~\cite{devries2017cutout}.
Datasets diversified with these augmented samples are shown to effectively improve the generalization performance of the trained model. Furthermore, data augmentation is proven to be effective not only for promoting generalization but also in boosting the robustness of a model~\cite{hendrycks2019augmix} and acquiring visual representations without human supervision~\cite{chen2020simple, moco}.

\begin{figure*}[t]
\begin{center}
\vspace{-0.1in}
\centerline{\includegraphics[width=0.72\textwidth]{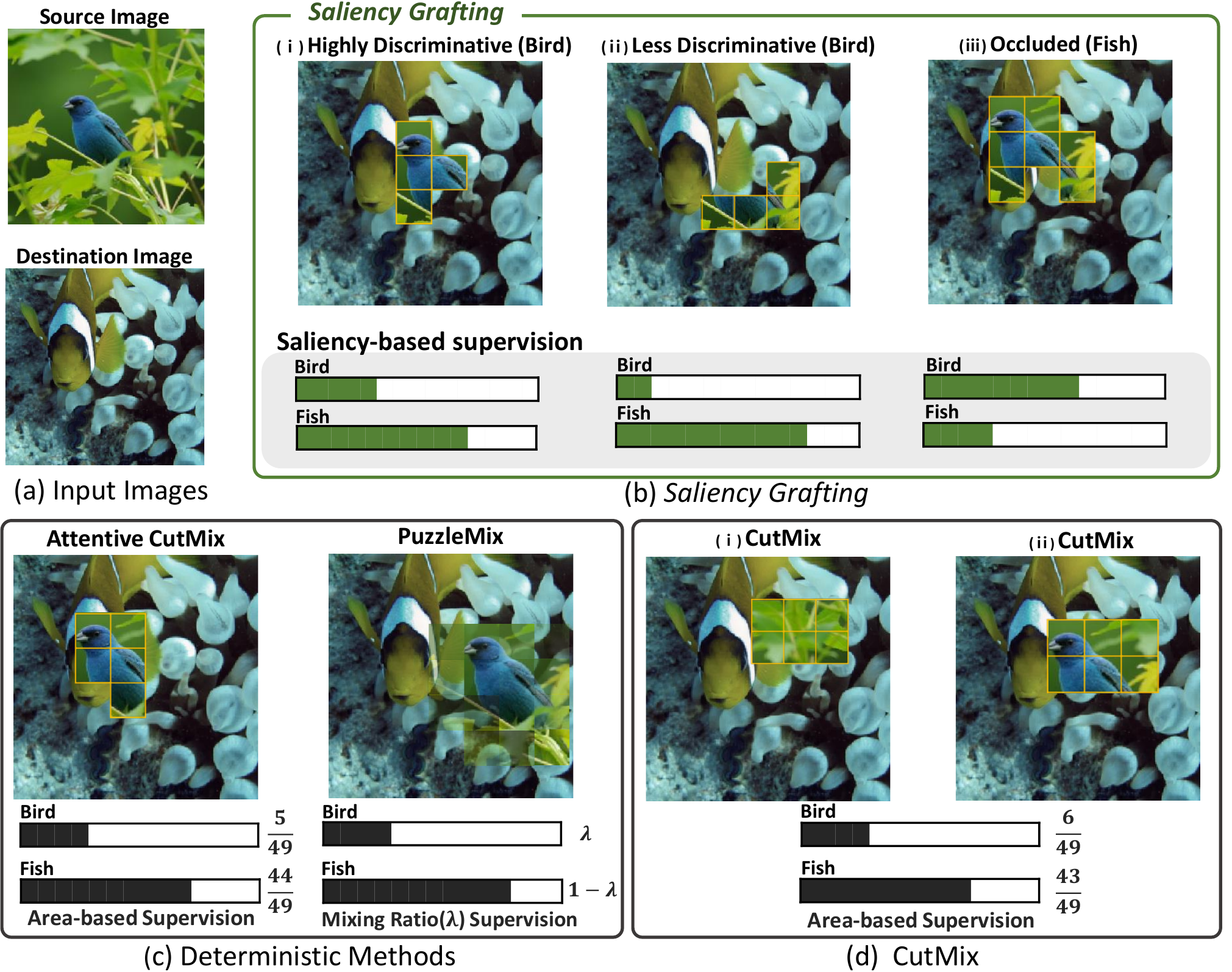}}
\caption{\small Comparison of augmented samples generated by mixup-based augmentations. (a) Source and destination images to be used in augmentation. (b) \emph{Saliency Grafting} produces diverse samples, including samples that do not contain the maximum saliency region. For all kinds of diverse samples, their labels are correctly rectified. (c) Deterministic saliency-based methods produce semantically plausible labels, but lack diversity since the maximum saliency region is always included. (d) CutMix generates diverse samples but produces misleading labels.}
\label{fig:sample_figure}
\end{center}
\vspace{-0.35in}
\end{figure*}

To this end, conventional augmentation methods have focused on creating new images by transforming a given image using means such as flipping, resizing, and more. However, a recently proposed augmentation method called Mixup~\cite{mixup} proposed the idea of crafting a new sample out of a pair of samples by taking a convex combination of them. Inspired by this pioneering work, \citet{yun2019cutmix} proposed CutMix, a progeny of Mixup and Cutout~\cite{devries2017cutout}, which crops a random region of an image and pasting it on another. These methods are able to generate a wider variety of samples while effectively compensating for the loss of information caused by actions such as cropping.
However, such context-agnostic nature of these methods gives way to creating samples that are potentially harmful. Since the images are combined randomly without considering their contexts and labels, incorrect augmentation is destined to occur (see Figure \ref{fig:sample_figure}(d)). For instance, an object can be cropped out and replaced by a different kind of object from another image, or the background part of the image can be pasted on top of an existing object. Even worse, their labels are naively mixed according only to their mixing proportions, disregarding any information transfer or loss caused by the data mixing. The harmfulness of semantically unaware label mixing was previously reported in \cite{adamixup}. This mismatch in data and its supervision signal yields harmful samples.   

To address this problem, saliency-guided augmentation methods have been recently proposed~\cite{attentivemix, puzzlemix, saliencymix}. These approaches allegedly refrain from generating harmful samples by preserving the region of maximum saliency based on the saliency maps of the image.
Attentive Cutmix~\cite{attentivemix} preserves the maximum saliency regions of the donor image by locating the $k$-most salient patches of it and merging them on top of the acceptor image. SaliencyMix~\cite{saliencymix} constructs a bounding box around the maximum saliency region and crops this box to place it on top of the acceptor image. PuzzleMix \cite{puzzlemix} tries to salvage the most salient regions of each image by mixing one to another and solving an optimal transport problem and region-wise mixup ratio to maximize the saliency of the created sample.
However, these precautionary measures sacrifice sample diversity - which is the advantage of previous CutMix-based methods. Unlike CutMix that teaches the model to attend to the whole object by probabilistically choosing diverse regions of the image, maximum saliency methods lose this feature as the most discriminative region is always included in the resulting image, biasing the model to depend on such regions.
Moreover, they still overlook making appropriate supervision to describe the augmented image properly, and use semantically inaccurate labels determined by the mixing ratio or the size of the pasted region, which can easily mislead the network (see Figure \ref{fig:sample_figure}(c)). 

To solve the drawbacks present in contemporary augmentation methods, we propose \textit{Saliency Grafting}, a novel data augmentation method that can generate \emph{diverse and innocuous} augmented data (see Figure \ref{fig:sample_figure}(b)). Instead of blindingly selecting the maximum saliency region, our method scales and thresholds the saliency map to grant all salient regions equal chance. The selected regions are then imposed with Bernoulli distribution and sampled to generate stochastic patches. These patches are then `grafted' on top of another image. Moreover, to compensate for the side effects of grafting such as label mismatch, we propose a novel label mixing strategy: saliency-guided label mixing. By mixing the labels of the two images according to their \textit{saliency} instead of their area, potential bad apples are effectively neutralized.

Our contribution is threefold:
\begin{itemize}
    
    \item We discuss the potential weaknesses of current Mixup-based augmentation strategies and present a novel data augmentation strategy that can generate diverse yet meaningful data through saliency-based sampling.
    
    \item We present a novel label mixing method to calibrate the generated label to match the information contained in the newly generated data.
    
    \item Through extensive experiments, we show that models trained with our method outperform others - even under data corruption or data scarcity.
\end{itemize}

\section{Related work}

\paragraph{Data augmentation}\label{section:relatedwork-saliency}

Image data augmentation played a formidable role in breakthroughs of deep learning based computer vision~\cite{lecun1998gradient, alexnet, vgg}. Recently, regional dropout methods such as Cutout~\cite{devries2017cutout}, Dropblock~\cite{ghiasi2018dropblock} and Random Erasing~\cite{zhong2020random} were proposed to promote generalization by removing selected regions of an image or a feature map to diversify the model's focus. However, the removed regions are bound to suffer from information loss.
The recently proposed Mixup~\cite{mixup} and its variants~\cite{manifold, adamixup}, shifted the augmentation paradigm by not only transforming a sample but using a pair of samples to create a new augmented sample via convex combination. Although successful on multiple domains, Mixup is met with lost opportunities when applied to images as it cannot exploit their spatial locality.
To remedy this issue, Cutmix~\cite{yun2019cutmix}, a method combining Cutout and Mixup, was proposed. By cropping out a region then filling it with a patch of another image, Cutmix executes regional dropout with less information loss.
However, in Cutmix, a new problem arises as the random cut-and-paste strategy incurs semantic information loss and label mismatch. To fix this issue, methods exploiting maximum saliency regions were proposed. Attentive Cutmix~\cite{attentivemix} selects the top-$k$ regions to cut and paste to another image. SaliencyMix~\cite{saliencymix} creates a bounding box around the maximum saliency region, and pastes the box on another image. Puzzlemix~\cite{puzzlemix} takes a slightly different approach, where it selects maximum saliency regions of the two images and solves a transportation problem to maximize the saliency of the mixed image. However, since the maximum saliency region is always pertained, the model is deprived of the opportunities to learn from challenging but beneficial samples present in CutMix. 


\paragraph{Saliency methods}

In neuroscience literature, \citet{koch1987shifts} first proposed saliency maps as a means for understanding the attention patterns of the human visual cortex. As contemporary CNNs bear close resemblance to the visual cortex, it is plausible to adapt this tool to observe the inner workings of CNNs. These saliency techniques inspired by human attention are divided into two groups: bottom-up (backward) and top-down (forward)~\cite{katsuki2014bottom}.
For backward methods, saliency is determined in a class-discriminative fashion. Starting from the output of the network, the saliency signal is back-propagated starting from the label logit and attributed to the regions of the input image. \cite{simonyan2013deep, zhou2016cam,gradcam} utilize the backpropagated gradients to construct saliency maps. Methods such as \cite{lrp, rap} proposed to backpropagate saliency scores with carefully designed backpropagation rules that preserve the total saliency score across a selected layer.
On the other hand, forward saliency techniques start from the input layer and accumulate the detected signals up the network. The accumulated signals are then extracted at a higher convolutional layer (often the last convolutional layer) to obtain a saliency map. Unlike backward approaches, forward methods are class-agnostic as the convolutional layers extract features from all possible objects inside an image to support the last classifier. These maps are used in a variety of fields such as classification~\cite{oquab2015object} and transfer learning~\cite{zagoruyko2017attentiontransfer}.

\section{Preliminaries}
\begin{table*}[t]
\caption{\small Overview of various mixed sample augmentations.}
\vspace{-0.1in}
\begin{center}
\begin{small}
\begin{tabular}{lcc}
\toprule
\textbf{Method} & \textbf{Augmentation function $\phi$} & \textbf{Label mixing function    $\psi$} \\
\hline
Mixup & $\lambda x_i + (1-\lambda) x_{j}$& \multirow{4}{*}{$ \lambda y_i + (1-\lambda) y_{j}$} \\
Manifold mixup &  $\lambda h(x_{i}) + (1-\lambda) h(x_{j})$  & \\
CutMix &  $ \1^\mathrm{Rect}_{i}  \odot x_{i} + (1-\1^\mathrm{Rect}_{i}) \odot x_{j}  $ & \\
Puzzle Mix & $ Z \odot \Pi^{T}_{i}x_{i} + (1-Z) \odot \Pi^{T}_{j}x_{j} $ & \\
\hline
Saliency Grafting &  $ M_{i} \odot x_{i} +  (1-M_{i}) \odot x_{j} $  & $\lambda(S_i, S_j, M_i) y_{i} + \big(1-\lambda(S_i, S_j, M_i)\big)y_j$\\
\bottomrule
\end{tabular}
\end{small}
\end{center}
\label{tb:method-eqns}
\end{table*}
We first clarify the notations used throughout the section by describing a general form of Mixup-based augmentation procedures.
Let $f_{\theta}(\cdot)$ be a Convolutional Neural Network (CNN) parametrized by $\theta$. For a given batch $B$ of input data $ \{ x_1, \hdots, x_m \} \in \mathcal{X}^{|B|}$ and the corresponding labels $ \{ y_1, \hdots, y_m\} \in \mathcal{Y}^{|B|}$, a mixed image $\tilde{x}$ is generated by the augmentation function $\phi(\cdot)$ and the corresponding label $\tilde{y}$ is created through the label mixing function $\psi(\cdot)$: $\tilde{x} = \phi(x_i,x_j)$ and $\tilde{y} = \psi(y_i, y_j)$ for data index $i$ and its random permutation $j$. 

Then, Mixup-based augmentation methods define their own $\phi(\cdot)$ as a pixel-wise convex combination of two randomly selected pair, as follows:
\begin{align}
    \phi(x_i, x_j) = M_\lambda \odot h(x_i) + (\1 - M_\lambda) \odot h(x_j)
\end{align}
where $M_\lambda$ is a mixing matrix controlled by a mixing ratio $\lambda$, $\odot$ is the element-wise Hadamard product, and $h(\cdot)$ is some pre-processing function.

The vanilla (input) Mixup defines the augmentation function $\phi$ as $\phi(x_i, x_j) = \lambda x_i + (1-\lambda) x_j$. Manifold Mixup uses similar function $\phi(x_i, x_j) = \lambda h(x_i) + (1-\lambda) h(x_j)$ but with the latent features. In CutMix, the augmentation function $\phi$ is defined as $ \1^\mathrm{Rect}_{i} \odot x_{i} +  (1-\1^\mathrm{Rect}_{i}) \odot x_{j}$. This method randomly cuts a rectangular region $\1^\mathrm{Rect}_{i}$ from the source image $x_{i}$  with area proportional to $\lambda$  and pastes it onto the destination image $x_{j}$. PuzzleMix, recent saliency-based Mixup variant, employs the augmentation function $ \phi(x_i, x_j) = Z^{*} \odot \Pi^{*T}_{i}x_{i} +  (\1-Z^{*}) \odot \Pi^{*T}_{j} x_j$. This method exploits the image transportation plan $\Pi$ and region-wise mask matrix $Z$ to maximize the saliency of the mixed image. Note that unlike the vanilla Mixup, $Z$ is discretized region-wise mixing matrix that satisfies $\lambda = \frac{1}{n} \sum_{s}\sum_{t}Z_{st}$ for given mixing ratio $\lambda$. 
To find the optimal transportation plan $\Pi^{*}$ and region-wise mask $Z^{*}$ for the maximum saliency, PuzzleMix solves additional optimization problems in an alternating fashion, per \emph{each} iteration. 

Although it is a simpler scalar function, the label function $\psi(\cdot)$ is also defined in a similar form to the augmentation function $\phi(\cdot)$:
\begin{align}
\psi({y_i, y_j}) = \rho y_i + (1-\rho) y_{j}
\end{align}
where $\rho$ is a label mixing coefficient determined by the sample pair $(x_i, y_i), (x_{j}, y_{j})$ and the mixing ratio $\lambda$ from $\phi$. However, in all methods mentioned above, this $\rho$ simply depends on $\lambda$, disregarding the contents of sample pair $x_i$ and $x_j$: $\tilde{y} = \lambda y_{i} + (1-\lambda) y_{j}$. 
Table~\ref{tb:method-eqns} summarizes $\phi$ and $\psi$ for the augmentation methods described above.   


\begin{figure*}[t]
\begin{center}
\vspace{-0.1in}
\centerline{\includegraphics[width=0.73\textwidth]{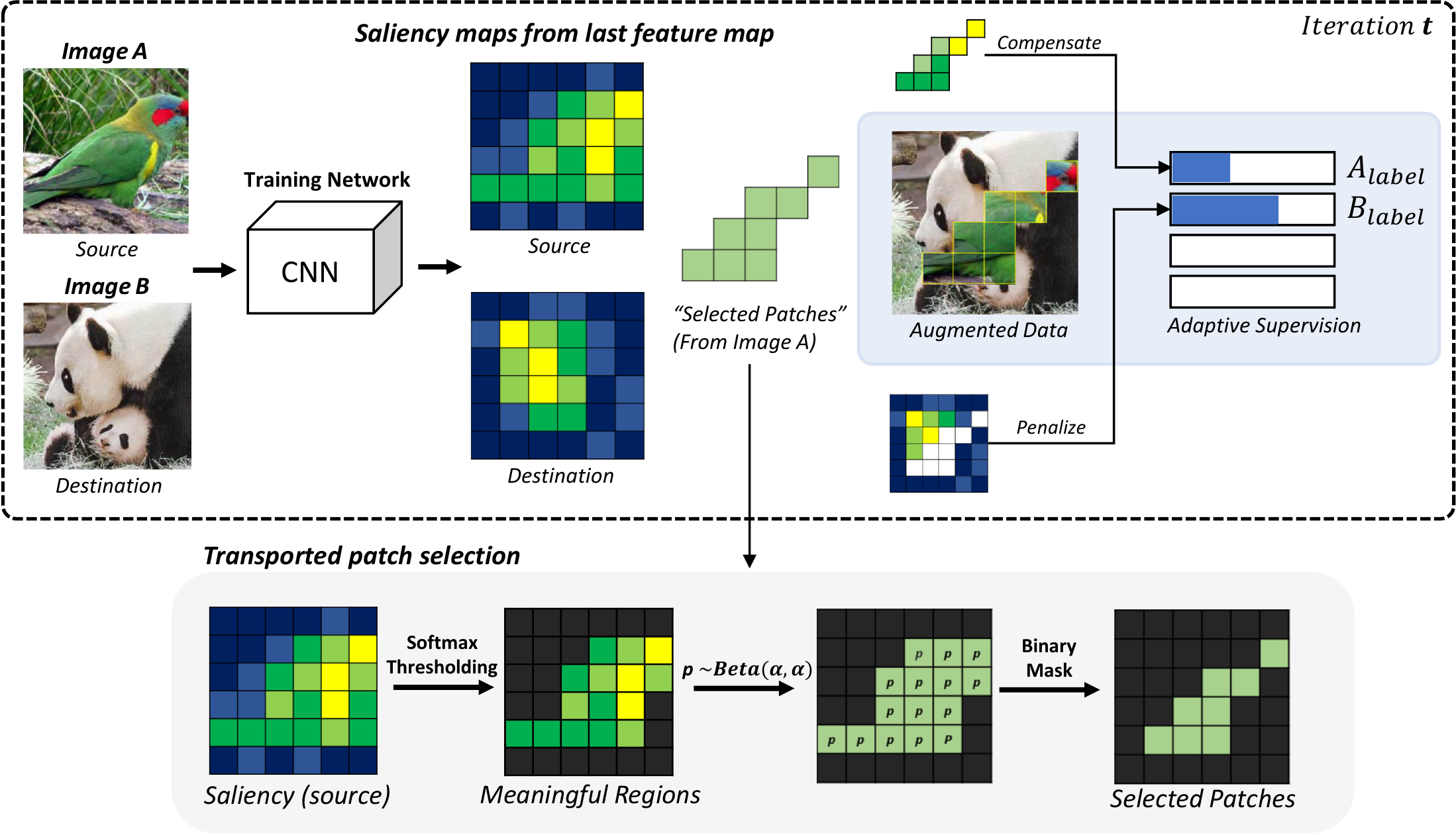}}
\vspace{-0.05in}
\caption{\small Overview of \textit{Saliency Grafting}. The source and destination images are drawn from the mini-batch and fed forward through the training network, producing their respective forward saliency maps. The source saliency map is thresholded and sampled using a region-wise i.i.d. Bernoulli distribution. Patches of the source image that corresponds to the resulting image are grafted to the destination image.}
\label{fig:main_figure}
\end{center}
\vspace{-0.35in}
\end{figure*}

\section{Saliency Grafting}
We now describe our simple approach, \emph{Saliency Grafting}, that creates diverse and innocuous Mixup augmentation based on the content of instances being merged. Two key innovations in \emph{Saliency Grafting} are stochastic patch selection (Section \ref{Subsec:PatchSelection}) and label mixing (Section \ref{Subsec:LabelMixing}), both of which utilize the saliency information at the core. Last but not least, another important element of \emph{Saliency Grafting} is choosing a saliency map generation method (Section \ref{Subsec:SaliencyMap}) for the above two main components while keeping the learning cost to a minimum. The overall procedure is described in Figure~\ref{fig:main_figure}. Now we discuss the details of each component in the subsequent subsections.

\subsection{Stochastic salient patch selection}\label{Subsec:PatchSelection}

The stochastic patch selection of \emph{Saliency Grafting} aims to choose regions that can create diverse and meaningful instances. The key question here is how to select regions to be grafted, given a saliency matrix $S_i$ for the source image $x_i$ (whose element $S_{st}$ indicates the saliency for a region $(s,t)$ of $x_i$). As in recent studies \cite{puzzlemix, attentivemix}, if only regions with high intensity of $S_{st}$ are always selected, then these regions - which are already easy to judge by the model - are continuously augmented in the iterative training procedure. As a result, the model is repeatedly exposed to the same grafting patch, which would iteratively amplify the model's attention on the selected regions and deprive the opportunity to learn how to attend to other parts and structures of the object.

In order to eliminate this \emph{selection bias}, the patch selection of \emph{Saliency Grafting} consists of two steps: i) softmax thresholding and ii) stochastic sampling.    

\paragraph{Softmax thresholding}

To neutralize the selection bias due to the intensity of saliency, we normalize the saliency map by applying the softmax function and then binarize the map with some threshold $\sigma$:
\begin{align}
S'_{st}(\bm{x} ; T) = \frac{\exp \left(S_{st}(\bm{x}) / T\right)}{\sum_{h}^{H} \sum_{w}^{W} \exp \left(S_{hw}(\bm{x}) / T\right)}, \\ 
S''_{st}(\bm{x} ; T) =
    \begin{cases}
        1,  \text{ if $S'_{st}(\bm{x} ; T)>\sigma$}\\
        0,  \text{ otherwise}
    \end{cases}
\end{align}
given the temperature hyperparameter $T$ to control the sharpness of the normalized saliency map. Here, threshold $\sigma$ has a variety of options, but we adopt the threshold as $\sigma_{\text{mean}} = \frac{1}{HW}\sum_{h}^{H} \sum_{w}^{W} S'_{hw}$, using the mean value of the normalized saliency map.

\paragraph{Stochastic sampling}
Although the selection bias is significantly mitigated by thresholding, the high intensity regions are never \textit{removed}, as the softmax function preserves the order of the regions. To address this issue, we stochastically sample the grafting regions based on the binarized saliency map produced above. The final mixing matrix $M_{i}$ is constructed by taking the Hadamard product of $S_{i}''$ and a region-wise i.i.d. random Bernoulli matrix of same dimensions $P \sim \mathbf{Bern}(p_{B})$:
$M_{i} = P \odot S''_{i}$.
Here, the batch-wise sampling probability $p_{B}$ is drawn from a Beta distribution $p_{B} \sim Beta(\alpha, \alpha)$. The final augmentation function $\phi$ for \emph{Saliency Grafting} is $M_{i} \odot x_{i} +  (1-M_{i}) \odot x_{j}$.

\subsection{Calibrated label mix based on saliency maps}\label{Subsec:LabelMixing}

In addition to the method of grafting diverse and innocuous augmentations described in the previous section, attaching an appropriate label for supervision to the generated data is also the core of \emph{Saliency Grafting}. Although extreme, to highlight the drawbacks of the existing label mixing strategy used in all baselines, suppose that source image $x_i$ is combined with destination image $x_j$, both of which have saliency concentrated in some small regions. Suppose further that this region of $x_i$ is selected and grafted to the region where the original class of destination $x_j$ is concentrated. Then, most of the information of class $y_i$ is retained while most of the information on class $y_j$ is lost. However, if the label is determined in proportion to the mixing rate or the size of the area used, as all the baselines do, the generated label will be close to class $y_j$ since most areas of it originally came from the destination image $x_j$.


To tackle this issue, we propose a novel label mixing procedure that can adaptively mix the labels again based on saliency maps. Regarding the destination image $x_{j}$ receiving the graft, the ground truth label $y_{j}$ is penalized according to the degree of occlusion. Specifically, the importance of the destination image $I({S_j, 1-M_i})$\footnote{We use $\ell_2$ norm to define the importance $I$ in the sense that the overall saliency is simply the same as the sum of saliency in each region, but similar importance can be obtained with other norms.} given the mixing matrix $M_i$ is calibrated using the saliency values of the remaining part not occluded by the source image, 
$I({S_j, 1-M_i}) = \frac{\|S_{j}\odot( \mathbf{1}- M_{i}) \|_{2}}{\|S_{j}\|_{2}}$.
On the other hand, with regard to the source image $x_{i}$ giving the graft, the corresponding label $y_{i}$ is compensated in proportion to the importance of the selected region: $I(S_{i}, M_{i}) = \frac{\|S_{i}\odot M_{i} \|_{2}}{\|S_{i}\|_{2}}$.

The final label mixing ratio is computed based on the relative importance of $x_i$ and $x_j$, so that their coefficients sum to $1$ to define the calibrated label mixing function $\psi$.
\begin{align}
\psi(y_i, y_j) = \lambda(S_i, S_j, M_i) y_{i} + \big(1-\lambda(S_i, S_j, M_i)\big) y_{j} \\
\text{where} \quad \lambda(S_i, S_j, M_i) = \frac{I(S_{i}, M_{i})}{I(S_{i}, M_{i})+I(S_{j}, 1-M_{i})} \nonumber
\end{align}


\subsection{Saliency map generation} \label{Subsec:SaliencyMap}


Technically, \emph{Saliency Grafting} can be combined with various saliency generation methods without the dependence on a specific method. However, the caveat here is that the performance of \emph{Saliency Grafting} is, by design, highly affected by the quality of the saliency map, or how accurately the saliency map corresponds to the ground truth label. From this point of view, the forward saliency methods, which 
incur less false negatives, may support \emph{Salient Grafting} more stably than the backward methods (see Section \ref{section:relatedwork-saliency} for forward and backward saliency methods). We also provide the performance comparison in Appendix A. This is because the backward methods are likely to break down and exclude true salient regions when the model fails to predict the true label, whereas the forward methods preserve all the feature maps inside the saliency map, i.e., they act like a class-agnostic saliency detector \cite{visualizing}. 

In an environment where there is no separate pre-trained model, another advantage of using forward saliency is gained: Saliency maps can be naturally constructed based on the terms already calculated in the learning process. In this environment, since the generated maps can be noisy in the early phases of training, we employ warmup epochs without data augmentation.



We now describe the choice of generating the saliency maps to guide our augmentation process. We adopt the channel-collapsed absolute feature map of the model as our saliency map, mainly due to its simplicity:
    $S^{(l)} = \sum_{c=1}^{C} \big|A_{c}^{(l)}\big|$
where $A \in \mathbb{R}^{C \times H \times W}$ is the feature map at the $l$-th layer. Albeit it is possible to extract saliency maps from any designated layer in the network, we extract the maps from the last convolutional layer as it generally conveys the high-level spatial information~\cite{bengio2013representation}. In practice, we randomly select the up/down-sampling scale of saliency maps per each mini-batch.

\section{Experiments}
We conduct a collection of experiments to test \emph{Saliency Grafting} against other baselines. First, we test the prediction performance on standard image classification datasets. Next, to confirm our claim that \emph{Saliency Grafting} can safely boost the diversity of augmented data, we design and conduct experiments to assess the sample diversity of each augmentation method. We also conduct multiple stress tests to measure the enhancement in generalization capability.
Finally, we perform an ablation study to investigate the contribution of each sub-component of \textit{Saliency Grafting}. Note that we train the models with both original and augmented images.

\subsection{Classification tasks}\label{cifar_cls}
\paragraph{CIFAR-100}
We evaluate our method \emph{Saliency Grafting} on CIFAR-100 dataset~\cite{cifar} using two neural networks: PyramidNet-200 with widening factor $\tilde{\alpha}=240$~\cite{pyramidnet} and WRN28-10~\cite{wideresnet}. For the PyramidNet-200, we follow the experimental setting of \citet{yun2019cutmix}, which trains PyramidNet-200 for 300 epochs. The baselines results on PyramidNet-200 are as reported in \citet{yun2019cutmix}. For WRN28-10, the network is trained for 400 epochs as following studies \cite{puzzlemix, manifold}. In this experiment, we reproduce other baselines following the original setting of each paper. Detailed settings are provided in Appendix B.
As shown in Table \ref{tb:pyramid} and Table \ref{tb:wrn_cifar}, \emph{Saliency Grafting} exhibits significant improvements for both architectures compared to other baselines. 
Furthermore, when used together with Shakedrop regularization~\cite{shakedrop}, \emph{Saliency Grafting} achieves additional enhancement - \textbf{13.05\%} Top-1 error. 

\begin{table}[t]
\caption{\small Error rates on CIFAR-100 for PyramidNet-200($\tilde{\alpha} = 240$) in comparison to state-of-the-art regularization methods. The experiment was performed three times and the averaged best error rates are reported.}
\small
\begin{center}
\vspace{-0.1in}
\begin{adjustbox}{width=0.9\linewidth}
\begin{tabular}{lcc}
\toprule
\textbf{PyramidNet-200} (\textbf{$\tilde{\alpha} = 240$)} & Top-1 & Top-5\\
(\# params: 26.8 M)& Error (\%) & Error (\%) \\
\hline
Vanilla & 16.45 & 3.69 \\
Cutout & 16.53 & 3.65 \\
DropBlock & 15.73 & 3.26 \\
Mixup ($\alpha=1.0$) & 15.63 & 3.99 \\
Manifold Mixup ($\alpha=1.0$) & 16.14 & 4.07 \\
ShakeDrop & 15.08 & 2.72 \\
Cutout + Mixup ($\alpha=1.0$) & 15.46 & 3.42 \\
Cutout + Manifold Mixup ($\alpha=1.0$) & 15.09 & 3.35 \\
CutMix & 14.47 & 2.97 \\
CutMix + ShakeDrop & 13.81 & 2.29 \\
Attentive CutMix (N = 6) & 15.24 \tiny $\pm 0.09$ & 3.46 \tiny$\pm 0.06$ \\
SaliencyMix & 14.74 \tiny $\pm 0.17$ & 3.07 \tiny $\pm 0.04$  \\
PuzzleMix & 14.78 & 3.08 \\
\textbf{Saliency Grafting} & \textbf{13.94} \tiny$\pm 0.11$  & \textbf{2.79} \tiny$\pm 0.09$ \\
\textbf{Saliency Grafting + ShakeDrop} & \textbf{13.05} \tiny$\pm 0.06$ & \textbf{2.18} \tiny$\pm 0.03$ \\
 
\bottomrule
\end{tabular}
\end{adjustbox}
\end{center}
\label{tb:pyramid}
\end{table}

 \begin{table}[t]
     \vspace{-0.1in}
\caption{\small Error rates on CIFAR-100 for WRN28-10 in comparison to data augmentation methods. The experiment was performed three times and the averaged best error rates with standard errors are reported. $\dagger$ indicates the reported result in the original paper.}
\begin{center}
\begin{small}
\begin{adjustbox}{width=0.75\linewidth}
\begin{tabular}{lcc}
\toprule
\textbf{WRN28-10}  & Top-1 & Top-5\\
(\# params: 36.5 M)& Error (\%) & Error (\%) \\
\hline
Vanilla & 20.74 \tiny{$\pm 0.06$} & 5.70 \tiny{$\pm 0.03$} \\
Mixup & 17.59 \tiny{$\pm 0.07$} & 5.18 \tiny{$\pm 0.14$}\\
Manifold Mixup$\dagger$  & 18.04 \tiny{$\pm 0.08$} & - \\
CutMix & 17.47 \tiny{$\pm 0.24$} & 4.80 \tiny{$\pm 0.42$} \\
AugMix & 19.19 \tiny{$\pm 0.04$} & 4.36 \tiny{$\pm 0.02$} \\
SaliencyMix & 16.38 \tiny{$\pm 0.04$} & 3.62 \tiny{$\pm 0.05$} \\
SaliencyMix (\small w/ dropout)$\dagger$ & 16.23 \tiny{$\pm 0.08$} & - \\
PuzzleMix & 16.00 \tiny{$\pm 0.03$} & 3.84 \tiny{$\pm 0.04$} \\
\textbf{Saliency Grafting} & \textbf{15.32} \tiny{$\pm 0.13$} & \textbf{3.54 }\tiny{$\pm 0.03$} \\

\bottomrule
\end{tabular}
\end{adjustbox}
\end{small}
\end{center}
\vskip -15pt
\label{tb:wrn_cifar}
\end{table}

\paragraph{Tiny-ImageNet}
We evaluate our method on another benchmark dataset - Tiny-ImageNet~\cite{tinyimagenet}. 
We train ResNet-18~\cite{resnet} for 600 epochs and report the converged error rates of the last 10 epochs, following one of Tiny-ImageNet experimental settings in \cite{puzzlemix}. Other data augmentation methods are evaluated using their author-released code and hyperparameters. Detailed experimental settings are described in Appendix B. The obtained results are shown in Table ~\ref{tb:resnet_tiny}. In line with the CIFAR-100 experiments, \emph{Saliency Grafting} consistently exhibits the best performance on this benchmark dataset.
\begin{table}[h]
\caption{\small Error rates on Tiny-ImageNet for ResNet-18 in comparison to data augmentations. The experiment was performed three times and the converged error rates with standard errors are reported.}
\vspace{-0.1cm}
\begin{center}
\begin{small}
\begin{adjustbox}{width=0.7\linewidth}
\begin{tabular}{lcc}
\toprule
\textbf{ResNet-18}  & Top-1 & Top-5\\
(\# params: 11.3 M)& Error (\%) & Error (\%) \\
\hline
Vanilla & 38.54 \tiny{$\pm 0.15$} & 18.53 \tiny{$\pm 0.11$} \\
Mixup & 37.37 \tiny{$\pm 0.14$} & 18.09 \tiny{$\pm 0.11$}\\
CutMix & 35.76 \tiny{$\pm 0.15$} & 15.82 \tiny{$\pm 0.20$} \\
SaliencyMix & 36.61 \tiny{$\pm 0.13$} & 16.31 \tiny{$\pm 0.27$} \\
PuzzleMix & 35.79\tiny{$\pm 0.17$} & 16.31 \tiny{$\pm 0.15$} \\
\textbf{Saliency Grafting} & \textbf{35.16} \tiny{$\pm 0.12$} & \textbf{15.02}\tiny{$\pm 0.07$} \\

\bottomrule
\end{tabular}
\end{adjustbox}
\end{small}
\end{center}
\vskip -15pt
\label{tb:resnet_tiny}
\end{table}

\paragraph{ImageNet}
For the ImageNet~\cite{imagenet} experiment, we train ResNet-50 for 100 epochs. We follow the training protocol in \citet{Wong2020Fast}, which includes cyclic learning rate, regularization on batch normalization layers, and mixed-precision training. This protocol also gradually resizes images during training, beginning with larger batches of smaller images and moving on to smaller batches of larger images later (\textit{image-resizing policy}). The baselines results are as reported in \cite{puzzlemix}. Detailed experimental settings are described in Appendix B. As shown in Table \ref{tb:imagenet}, \emph{Saliency Grafting} achieves again the best performance in both Top-1/Top-5 error rates. We confirm that ours can bring further performance improvement without image-resizing scheduling.

\begin{table}[h]
\vspace{-0.1 cm}
\caption{\small Comparison of state-of-the-art data augmentation methods on ImageNet dataset.}
\vspace{-0.35 cm}
\small
\begin{center}
\begin{adjustbox}{width=0.86\linewidth}
\begin{tabular}{lcc}
\toprule
\textbf{ResNet-50} & Top-1 & Top-5\\
(\# params: 25.6M)& Error (\%) & Error (\%) \\
\hline
Vanilla & 24.31 & 7.34 \\
Mixup & 22.99 & 6.48 \\
Manifold Mixup & 23.15 & 6.50 \\
CutMix & 22.92 & 6.55 \\
AugMix & 23.25 & 6.70 \\
PuzzleMix & 22.49 & 6.24 \\
\textbf{Saliency Grafting} & 22.35 & \textbf{6.19} \\
\textbf{Saliency Grafting \textit{\begin{footnotesize}(w/o image-resizing policy)\end{footnotesize}}}  & \textbf{22.26} & 6.29 \\

\bottomrule
\end{tabular}
\end{adjustbox}
\end{center}

\vspace{-0.2in}
\label{tb:imagenet}
\end{table}

\paragraph{Additional experiments} Due to the space constraint, three additional experiments are deferred to Appendix A. The first experiment shows that \emph{Saliency Grafting} is useful for \textbf{speech} dataset beyond the image classification task, and the second experiment (\textbf{weakly supervised object localization}) implies that the final model learned through \emph{Saliency Grafting} contains more useful saliency information. 

\subsection{Sample diversity}~\label{subsec:sample_diversity}
\begin{figure}[h]
\begin{center}
\centering
\includegraphics[width=0.664\linewidth]{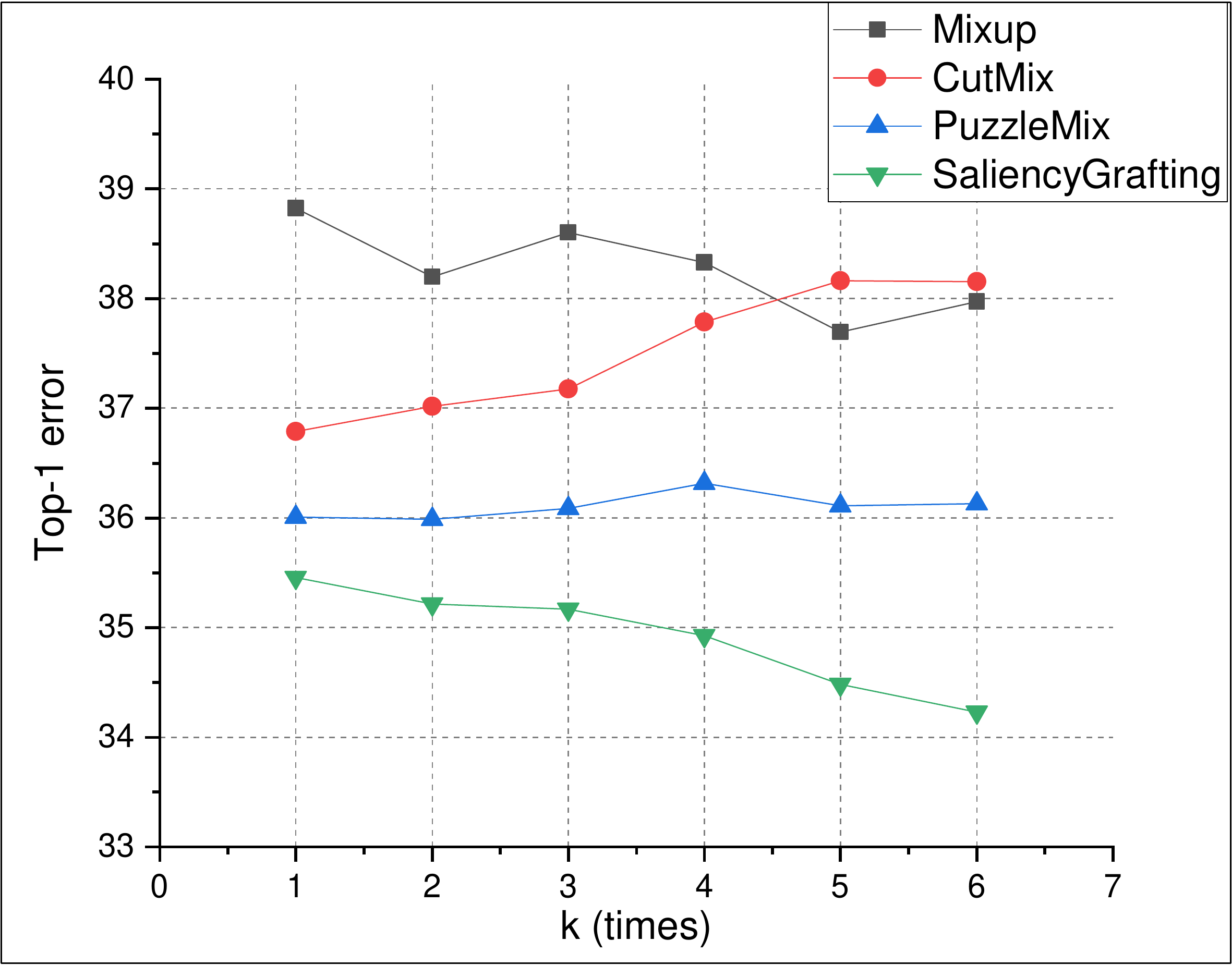}
\caption{\small Comparison of sample diversity by generating \textit{k}-times augmented data from a mini-batch. The experiment was performed five times and the averaged best error rates are reported.}
\label{fig:batch_k}
\end{center}
\vspace{-0.25in}
\end{figure}

\vspace{-0.15 in}
\paragraph{Generating augmented data \textit{k}-times}
We design intuitive experiment to compare \emph{Saliency Grafting} and other augmentation methods in terms of sample diversity. For every iteration, each method trains the network by generating additional augmented data \textit{k} times from the mini-batch; each method tries to diversify the mini-batch by producing \textit{k} independent augmented batches with its own randomness. To ensure sufficient diversity, the mixing ratio $\lambda$ is newly sampled for each augmented data. While varying \textit{k} from 1 to 6, we evaluate whether each method can obtain the performance gain due to sample diversity. We train the WRN28-10 for 200 epochs and use 20\% of the CIFAR-100 dataset to better confirm the diversity effect of the augmented data. In Figure~\ref{fig:batch_k}, the performance of \emph{Saliency Grafting} consistently improves as \textit{k} increases, whereas PuzzleMix, one of the representative maximum saliency strategies, does not show any gain in performance even when \textit{k} increases. In this sense, we believe that this is the direct evidence that generating multiple augmented instances by sampling the random mixing ratio is insufficient to ensure sample diversity in the case of maximum saliency approaches~\cite{puzzlemix,attentivemix}. However, since \emph{Saliency Grafting} exploits temperature-scaled thresholding with stochastic sampling, the model easily attends to the entire object as $k$ increases. It is also possible to properly supervise the augmented data through calibrated label mixing. Hence, the sample diversity can be guaranteed innocuity.

\subsection{Stress testing}

\paragraph{Data scarcity}
The situation where data augmentation is most needed is when data is scarce. In this condition, it is important to improve the generalization performance by increasing the data volume while preventing overfitting. To this end, we test our method against data scarcity by reducing the number of data per class to 50\%, 20\%, and 10\%, with the WRN28-10 model on the CIFAR-100 dataset. 
In Table~\ref{tb:scarcity}, \emph{Saliency Grafting} exhibits the best performance in every condition. 
These results are in line with the fact that, as investigated by \citet{rolnick2017deep}, corrupted labels severely degrade the performance when data is scarce. Since our method exploits adaptive label mixing to reduce the mismatch between data and labels while maintaining the diversity to prevent overfitting, the generalization performance can be enhanced even in extreme data scarcity conditions.


\begin{table}[h]
\vspace{-0.1cm}
\caption{\small Top-1 error rates on the CIFAR-100 with reduced number of data per class. The experiment was performed three times.} 
\vspace{-0.25cm}
\label{tb:scarcity}
\begin{center}
\scalebox{0.8}{
\begin{tabular} {lccc}
\toprule
\textbf{\# of data per class} & 50 (10 \%) & 100 (20 \%) & 250 (50 \%) \\
\hline
Vanilla & 59.96 \tiny{$\pm 0.50$} & 44.29 \tiny{$\pm 0.51$} & 31.19 \tiny{$\pm 0.20$}  \\
Mixup & 51.29 \tiny{$\pm 0.27$} & 38.80 \tiny{$\pm 0.51$} & 27.11 \tiny{$\pm 0.10$}  \\
CutMix & 52.90 \tiny{$\pm 0.18$} & 38.76 \tiny{$\pm 0.15$} & 26.58 \tiny{$\pm 0.10$} \\
PuzzleMix & 54.69 \tiny{$\pm 0.54$} & 38.66 \tiny{$\pm 0.16$} & 26.69 \tiny{$\pm 0.05$}  \\
\textbf{Saliency Grafting} & \textbf{51.01} \tiny{$\pm 0.31$} & \textbf{37.35} \tiny{$\pm 0.21$} & \textbf{25.56} \tiny{$\pm 0.24$}   \\
\bottomrule
\end{tabular}
}
\end{center}
\vskip -15pt
\end{table}

\paragraph{Partial occlusion of salient regions}
We demonstrate how existing saliency-guided augmentations fail to diversify the given data (Section~\ref{subsec:sample_diversity}). These methods are designed to \emph{always} preserve the maximum saliency region, but this strategy harms generalization by training the model to `expect' such a region (injecting bias), which is \emph{not} the case outside the lab where objects can be partially occluded. This forfeits the diversification effects of their crop-and-mix strategy, degrading performance.
To expose the dataset bias induced by previous saliency augmentations, we conduct an `occlusion experiment', where we remove the top-$k$ salient regions from the images then evaluate. Table~\ref{tb:occlusion} shows that as the occluded area gets larger, other methods perform worse than ours due to bias injection, while \textit{Saliency Grafting} scores the highest as stochastic sampling removes the bias.

\begin{table}[h]
\begin{center}
\vspace{-0.1 in}
\caption{\small Top-1 error rates on TinyImageNet with top-$k \%$ salient regions removed.}
\label{tb:occlusion}
\vspace{-0.1 in}
\begin{small}
\setlength{\tabcolsep}{1.5pt} 
\renewcommand{\arraystretch}{1.0} 
\scalebox{0.8}{
\begin{tabular}{@{\extracolsep{2pt}}lccc@{}}
\toprule

\textbf{Method} & \multicolumn{3}{c}{\textbf{Top-1 Error (\%)}} \\
\cline{2-4}
(ResNet-18) & $k=0\%$ & $k=12.5\%$ & $k=25\%$  \\
\cline{1-4}
                    SaliencyMix & 36.61 & 44.73 & 55.89   \\
                    PuzzleMix & 35.79 & 50.91 & 66.23     \\
                    \cline{1-4}
                    \textbf{SaliencyGrafting} & \textbf{35.16} & \textbf{42.98} & \textbf{52.19}   \\
\bottomrule
\end{tabular}
}
\vspace{-0.18 in}
\end{small}
\end{center}
\end{table}

\subsection{Ablation study} \label{Subsec:ablation}

\paragraph{Stochastic selection VS deterministic selection}
In Section~\ref{Subsec:PatchSelection}, we argued that the deterministic region selection process of existing maximum saliency methods~\cite{attentivemix, puzzlemix, saliencymix} leads to performance degradation. This was partly shown in Table~\ref{tb:pyramid}, \ref{tb:resnet_tiny}, and \ref{tb:imagenet} where such methods perform worse than CutMix. Here, we directly study the contribution of stochastic selection. We measure the classification accuracy on CIFAR-100 with two architectures where the deterministic top-$k$ selection of Attentive CutMix~\cite{attentivemix} is replaced by our stochastic selection. For fair comparison, the softmax temperature $T$ is adjusted to satisfy $\mathbb{E}_{i}[\sum_{s}\sum_{t} {M_{i, st}}] = k$. Results show that stochastic selection indeed outperforms deterministic selection (Table \ref{tb:ablation-pyr}).


\paragraph{Effect of threshold $\sigma$}

\begin{figure}[h]
\begin{center}
\centering
\vspace{-0.1in}
\includegraphics[width=0.72\linewidth]{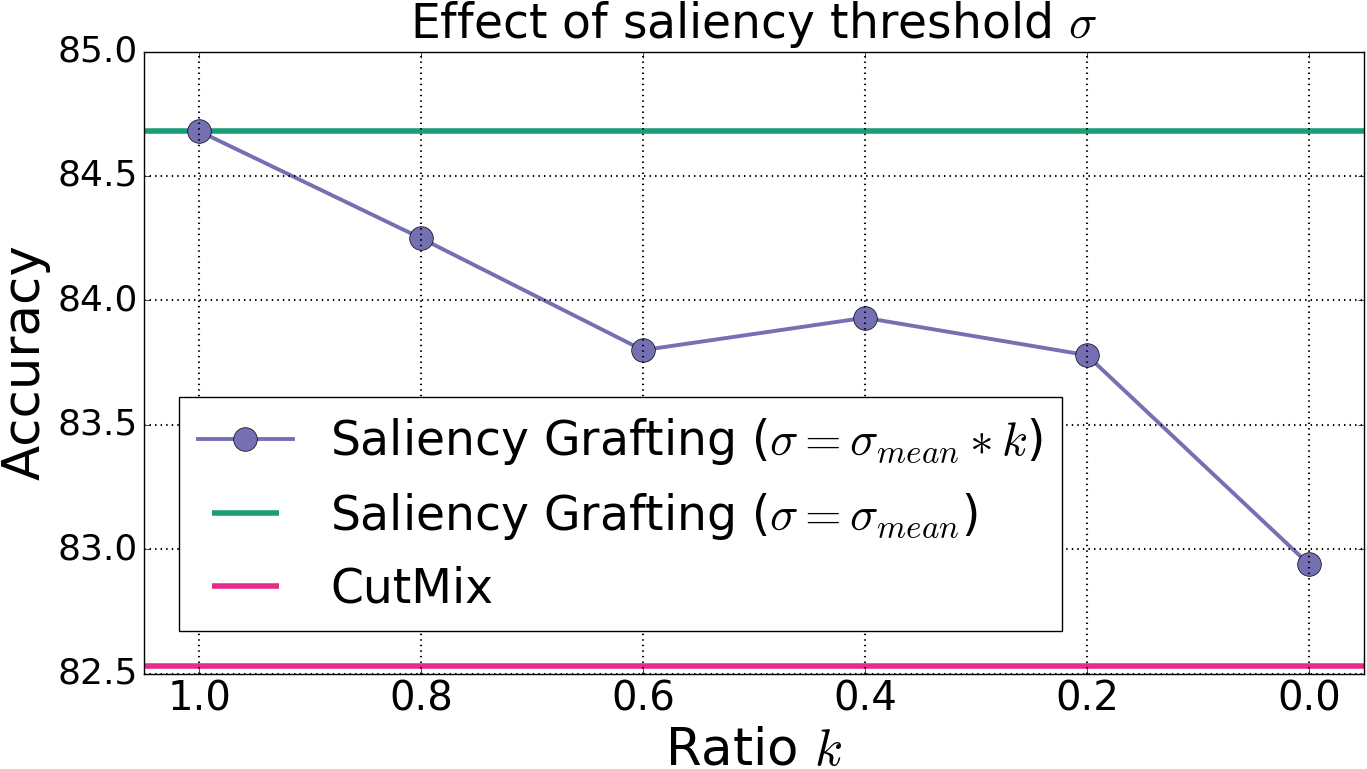}
\vspace{-0.1in}
\caption{\small Effect of threshold on CIFAR-100 with WRN28-10.}
\vspace{-0.2in}
\label{fig:noisy_saliency}
\end{center}
\end{figure}

Here, we conduct an experiment where we vary the saliency threshold $\sigma$. Figure~\ref{fig:noisy_saliency} shows that as we lower $\sigma$ below the normalized saliency mean $\sigma_{\text{mean}}$, non-salient regions are introduced, and the performance degenerates. At $\sigma = 0$, SG becomes saliency-agnostic~(which is near-equivalent to the CutMix strategy), and the performance of SG converges to the vicinity of CutMix.

\paragraph{Label mixing strategies}
In Section~\ref{Subsec:LabelMixing}, we discussed the pitfalls of naive area-based label mixing and proposed saliency-based label mixing as a solution. Here, we compare the two strategies. We experiment on CIFAR-100 with two architectures and replace the mixing strategy of \textit{Saliency Grafting} with area-based mixing. Results in Table~\ref{tb:ablation-pyr} confirm that saliency-based mixing outperforms area-based mixing.

\begin{table}[h]
\vspace{-0.05in}
\caption{\small Top-1 error rates on CIFAR-100 for PyramidNet-200 ($\tilde{\alpha} = 240$) and WRN28-10.}
\vspace{-0.25 cm}
\label{tb:ablation-pyr}
\begin{center}
\begin{small}
\scriptsize
\setlength{\tabcolsep}{1.30pt} 
\renewcommand{\arraystretch}{1.0} 
\begin{tabular}{@{\extracolsep{2pt}}lcc@{}}
\toprule
\textbf{Method} & \textbf{WRN28-10} & \textbf{PyramidNet-200}  \\
\cline{2-3}
 Saliency Grafting (SG) & Top-1 Error (\%) &  Top-1 Error (\%) \\
\hline
Deterministic + area labels & 16.34  & 14.63  \\
Stochastic + area labels  & 15.67 & 14.14 \\
Stochastic + saliency labels &  \textbf{15.32} & \textbf{13.94} \\
\bottomrule
\end{tabular}
\end{small}
\end{center}
\label{tb:novelty_ablation}
\vspace{-0.2 in}
\end{table}




\vspace{-0.01 in}
\section{Conclusion}
We have presented \textit{Saliency Grafting}, a data augmentation method that generates diverse saliency-guided samples via stochastic sampling and neutralizing any induced data-label mismatch with saliency-based label mixing. Through extensive experiments, we have shown that models equipped with \textit{Saliency Grafting} outperform existing mixup-based data augmentation techniques under both normal and extreme conditions while using less computational resources.

\section{Acknowledgements}
This work was supported by Institute of Information \& communications Technology Planning \& Evaluation (IITP) grant funded by the Korea government(MSIT) (No.2019-0-00075, Artificial Intelligence Graduate School Program(KAIST)) and National Research Foundation of Korea (NRF) grants (2019R1C1C1009192). This work was also partly supported by KAIST-NAVER Hypercreative AI Center.

\bibstyle{aaai22}
\bibliography{aaai22}

\clearpage

\appendix
\section{Additional experiments}\label{appendix:additionalEx}
\paragraph{Speech data}

To test our method on data outside the distribution of natural images, we evaluate our method on the Google Speech Commands dataset~\cite{speechcommand}. The training samples are first augmented in the time domain by applying random changes in amplitude, speed and pitch and in the frequency domain by stretching and time-shifting the spectrogram. Then, random background noises clip drawn from the noise compilation in the dataset are added to the samples. Finally, the samples are transformed into 32 $\times$ 32 mel-spectrograms by using 32 MFCC filters. To evaluate our method on this data, we use the WRN28-10 architecture. As shown in Table~\ref{tb:wrn_speech}, our method is able to outperform other methods in a non-natural image domain.

\begin{table}[h]

\caption{Top-1 error rates on Google Speech Commands in comparison to other augmentation methods. The experiment was performed three times and the test error rates with standard errors are reported.}
\begin{center}
\begin{small}
\begin{tabular}{lc}
\toprule
\textbf{WRN28-10}  & Top-1 \\
(\# params: 36.5 M)& Error (\%) \\
\hline
Vanilla & 2.81  \tiny{$\pm 0.04$} \\
Mixup & 2.72  \tiny{$\pm 0.02$} \\
CutMix & 2.62  \tiny{$\pm 0.03$} \\
\textbf{Saliency Grafting} & \textbf{2.51} \tiny{$\pm 0.04$} \\

\bottomrule
\end{tabular}
\end{small}
\end{center}
\vskip -8pt
\label{tb:wrn_speech}
\end{table}

\paragraph{Weakly supervised object localization}
To examine how our method affects the \textit{backward} saliency of a model (how a model `thinks'), we measure the weakly supervised object localization performance on the CUB200-2011 dataset~\cite{wah2011caltech}. We follow the experiment protocol of \citet{yun2019cutmix} except for the use of an ImageNet pretrained network. For ResNet-50, we slightly modify the last convolution layer to increase the feature map size from 7x7 to 14x14. We first obtain the backward saliency map with CAM~\cite{zhou2016cam}. The map is then thresholded using 15\% of the maximum value of CAM and enclosed by the smallest possible bounding box. We measure the Intersection-over-Union(IoU) between this estimated bounding box and the ground truth bounding box. For the localization of a single image to be correct, the IoU between the estimated bounding box and the ground truth box should be greater than 0.5, and simultaneously, the predicted class label should be correct. We use the Adam optimizer, and the initial learning rate, weight decay, batch size were 0.001,0.0001, and 32. The initial learning rate of the last fully-connected layers are set to 0.01. The learning rate is decaying by the factor of 0.1 per 150 epochs. All the experiments were performed three times and the averaged localization accuracies are reported.

\begin{table}[h]

\caption{\small Performance of weakly supervised object localization on the CUB200-2011 dataset.}
\begin{center}
\begin{small}
\begin{tabular}{lc}
\toprule
\textbf{Method} & Loc Acc(\%) \\

\hline
ResNet-50 + CAM &  29.09 \tiny{$\pm 1.21$} \\
ResNet-50 + Mixup &  32.94 \tiny{$\pm 0.11$} \\
ResNet-50 + CutMix & 27.95 \tiny{$\pm 0.52$}  \\
ResNet-50 + PuzzleMix & 35.43 \tiny{$\pm 0.49$}   \\
\textbf{ResNet-50 + Saliency Grafting} & \textbf{38.58} \tiny{$\pm 0.34$} \\
\bottomrule
\end{tabular}
\end{small}
\end{center}
\vskip -4pt
\label{tb:wsol}
\end{table}

\paragraph{Forward saliency VS Backward saliency}
To support our choice of forward saliency maps in Section~4.3 of the main paper, we conduct an additional experiment on CIFAR-100 with WRN28-10 where the forward saliency map of \textit{Saliency Grafting} is replaced by CAM~\cite{zhou2016cam}, a backward saliency map. The detailed settings are kept identical to Section~5.1 of the main paper~(refer to Appendix~\ref{appendix:cifar100}). Results show that the classification error increases when a backward saliency map is used (Table~\ref{tb:ablation-wrn}).

\begin{table}[h]
    \vspace{-0.2cm}
    \caption{\small Top-1/Top-5 error rates on CIFAR-100 for WRN28-10.}
    \vspace{-0.2 cm}
    \label{tb:ablation-wrn}
    \begin{center}
    \scalebox{0.8}{
        \begin{tabular}{lcc}
        \toprule
        \textbf{Method} & Top-1 & Top-5\\
        & Error (\%) & Error (\%) \\
        \hline
        Backward (CAM) & 15.70 & 3.8 \\
        Forward (ours)& \textbf{15.32} & \textbf{3.54} \\
        \bottomrule
        \end{tabular}
        }
    \end{center}
\vspace{-0.2cm}
\end{table}

\begin{figure}[h]
    \vspace{-0.1in}
\begin{center}
\centerline{\includegraphics[width=1.\linewidth]{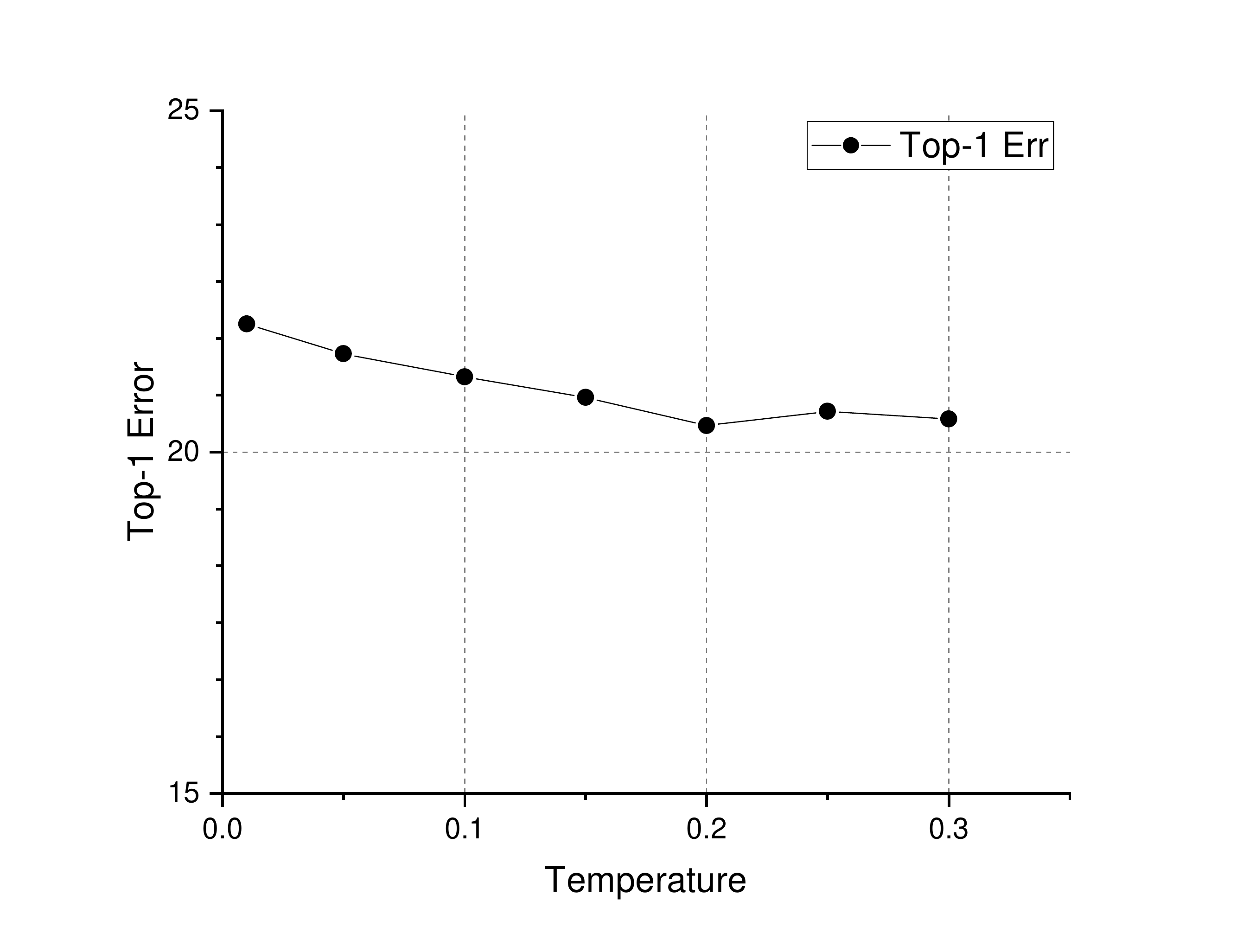}}
\vspace{-0.1in}
\caption{\emph{Saliency Grafting}'s sensitivity to temperature $T$.}
\label{fig:temperature_sensitivity}
\end{center}
\vspace{-0.2in}
\end{figure}

\paragraph{Sensitivity to temperature $\textbf{T}$}
\vskip -4pt
The threshold value of our method is determined by the mean value of the temperature-scaled saliency map. Note that the number of saliency regions greater than the expectation depends on the temperature $T$. As $T$ decreases, the softmax distribution becomes sharper and the number of saliency regions above the expectation decreases. That is, the mixing regions are selected from a smaller range. On the other hand, as $T$ increases, the distribution flattens so that nearly half the numbers are above the threshold. To see the sensitivity of model performance with respect to the softmax temperature, we conducted an additional experiment on the CIFAR-100 dataset with ResNet-18 by increasing the temperature from 0.01 to 0.30 (Figure~\ref{fig:temperature_sensitivity}). If we set a very small $T$, such as 0.01, only a small number of regions are mixed, resulting in a relatively small performance improvement. As we raise the temperature, the number of participating regions increases, resulting in a major increase in performance. When the temperature is sufficiently high, enough number of regions can participate in the mix. Thus, further increasing the temperature plateaus the performance.

\section{Detailed experimental settings} \label{appendix:exp-settings}

\subsection{CIFAR-100 Classification} \label{appendix:cifar100}
 We use stochastic gradient descent (SGD) with a momentum of 0.9 for both network models. For each mini-batch, mixing ratio $\lambda$ is sampled from \textit{Beta}(1,1) with regard to Mixup and CutMix. In the Manifold Mixup case, we adopt \textit{Beta}(2,2) for sampling distribution to follow the original paper. PuzzleMix has four hyperparameters: label smoothness term $\beta$, data smoothness term $\gamma$,  prior term $\eta$, and transport cost $\xi$. We use $(\beta, \gamma, \eta, \xi) = (1.2, 0.5, 0.2, 0.8)$. For the classification task, our method uses a temperature $T = 0.2$ and \textit{Beta}(2,2) for stochastic sampling. For early convergence, we warm up the model for 5 epochs. The described weight decay of each augmentation method is different for the CIFAR dataset, so the results of our paper are reported as having better results among 0.0005 and 0.0001. For the PyramidNet-200 network, the initial learning rate is set to 0.25 and decayed by the factor of 0.1 at 150 and 225 epoch. For the WRN28-10 network, the initial learning rate is set to 0.2 and decayed by the factor of 0.1 at the 200 and 300 epoch. All the experiments were performed three times with two TITAN XP GPUs and the averaged best error rates are reported.

\subsection{Tiny-ImageNet Classification} \label{appendix:tiny-imagenet}
For Tiny-ImageNet, we train the ResNet-18 model for 600 epochs using images resized to 64 $\times$ 64. As in the CIFAR-100 experiments, we use a temperature $T = 0.2$ and \textit{Beta}(2,2) for stochastic sampling. We randomly down-sample the resolution of saliency map to one of \{4$\times$4,8$\times$8\} to support multi-scale saliency. We also warm up the model for 5 epochs. Other data augmentation baselines are evaluated using the authors' hyperparameters as described above Section~\ref{appendix:cifar100}. We use the SGD optimizer with momentum 0.9 and weight decay 0.0001. The initial learning rate is set to 0.2 and decayed by the factor of 0.1 at 300 and 450 epoch. All the experiments were performed three times with one TITAN XP GPU and the converged error rates with standard errors for the last 10 epochs are reported as \cite{puzzlemix}.
 
\subsection{ImageNet Classification}\label{appendix:imagenet}
For the ImageNet, we follow the training process in \cite{Wong2020Fast,puzzlemix}. We train the ResNet-50 model for 100 epochs with 4 RTX TITAN GPUs. Specifically, we use cyclic learning rate scheduling, mixed-precision training, and weight decay regularization from batch normalization layers. Moreover, this protocol progressively resizes images during the training phase. We test our method on both settings (w/ image-resizing policy, w/o image-resizing policy). Our method adopts temperature $T = 0.3$ and sampling probabilities are sampled from $Beta(2,2)$. We randomly up/down-sample the resolution of saliency map to one of \{4$\times$4, 7$\times$7, 8$\times$8\}. We use the SGD optimizer, and the initial learning rate, momentum, and weight decay are 0.1, 0.9, 0.0001. For the fixed-size setting (w/o image-resizing policy), the images are fixed at $224 \times 224$ and the batch size is 256. we warm up the model for 5 epochs.

\clearpage
\onecolumn
\section{Examples}\label{appendix:examples}

\begin{figure*}[hbt!]
\begin{center}
\centerline{\includegraphics[width=0.81\textwidth]{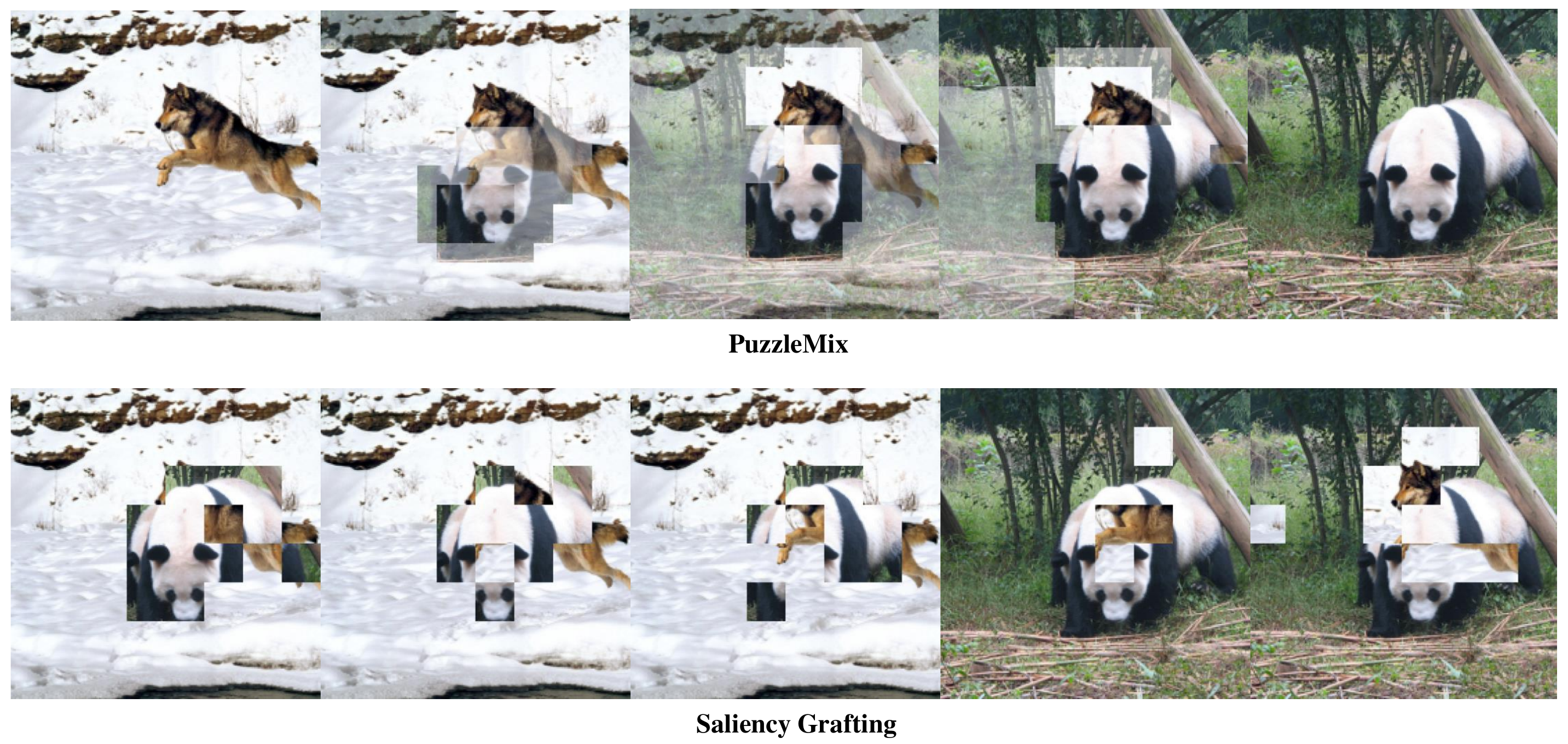}}
\centerline{\includegraphics[width=0.81\textwidth]{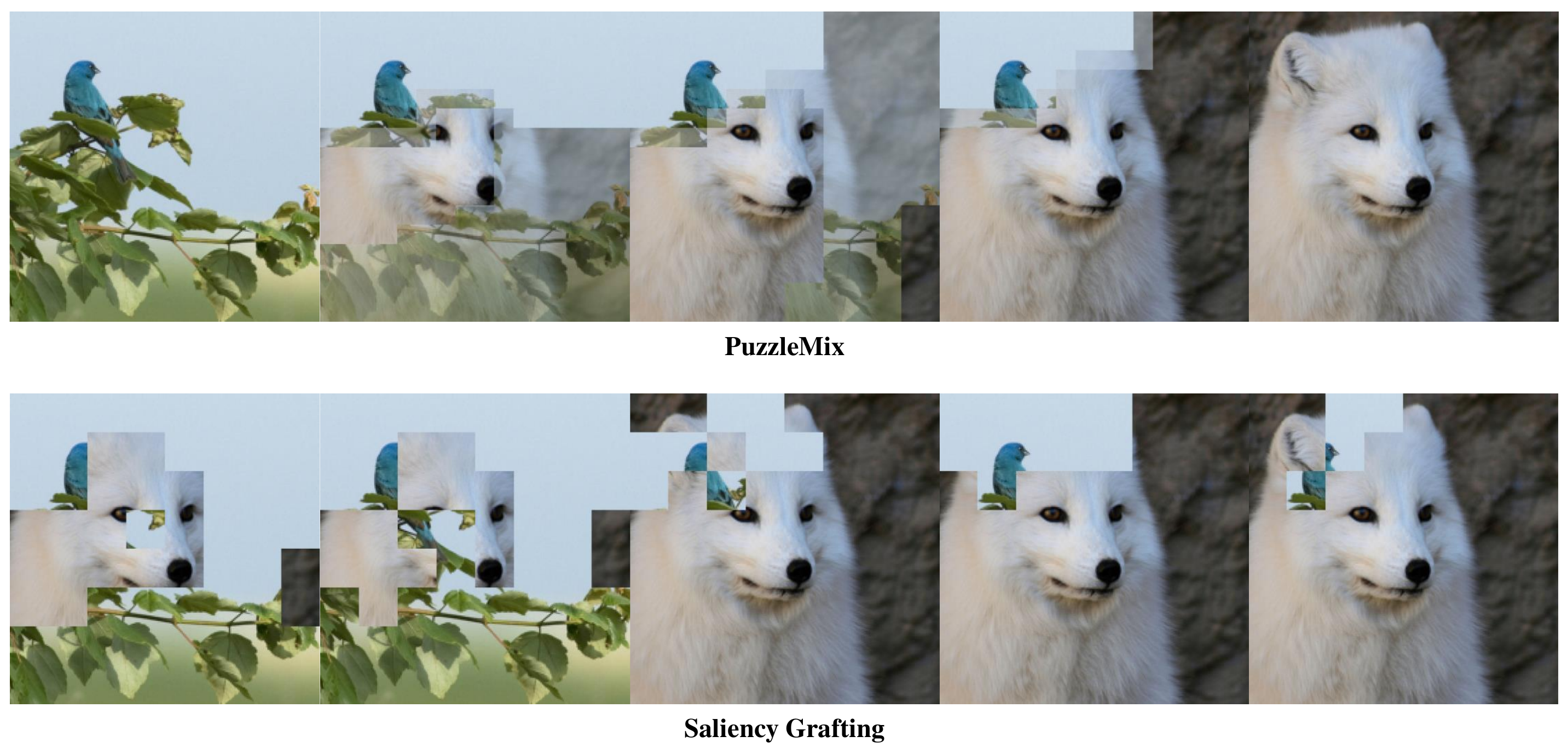}}
\centerline{\includegraphics[width=0.81\textwidth]{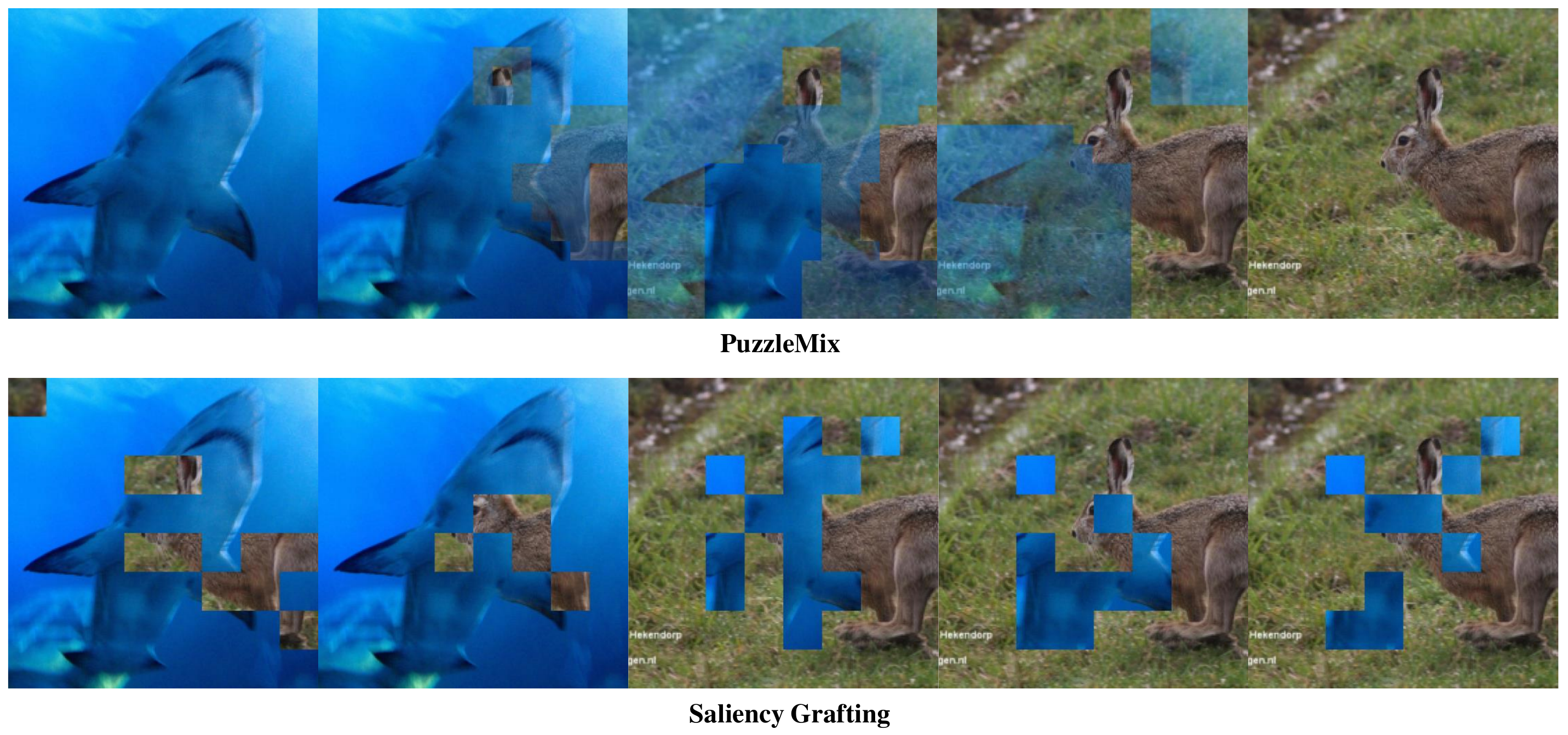}}
\caption{Comparison of diversity between \emph{Saliency Grafting} and PuzzleMix images.}
\label{fig:diversity_comparison}
\end{center}
\end{figure*}

\end{document}